%% file: acl_latex.tex
\ifdefined\pdfoutput\pdfoutput=1\fi

\documentclass[11pt]{article}

\usepackage[preprint]{acl}

\usepackage{times}
\usepackage{latexsym}
\usepackage{makecell} 

\usepackage[T1]{fontenc}

\usepackage[utf8]{inputenc}

\usepackage{microtype}

\usepackage{inconsolata}

\usepackage{enumerate}
\usepackage{amsthm}
\usepackage{graphicx}
\usepackage{epstopdf}
\usepackage{caption}
\usepackage{color}
\usepackage{enumitem}
\setlist[enumerate]{nosep} 
\usepackage{array}
\usepackage{tabularx}
\usepackage{multirow}
\usepackage{bm}
\usepackage{booktabs}
\usepackage{amssymb}
\usepackage{tabularx}
\newcolumntype{Y}{>{\centering\arraybackslash}X}

\usepackage{tcolorbox}
\tcbuselibrary{breakable}
\tcbuselibrary{skins}
\usepackage{caption}
\usepackage{subcaption}
\usepackage{listings}
\usepackage{wrapfig}
\usepackage{bbding}
\tcbuselibrary{skins}
\usepackage{colortbl}
\usepackage{fontawesome5}
\usepackage{authblk}
\usepackage{amsmath}
\usepackage{hyperref}
\usepackage{etoolbox}
\preto{\abstractkeywords}{\nolinenumbers} 

\title{Speculative Pipeline Decoding: Higher-Accuracy Drafting with Hidden Latency via Pipeline Parallelism}

\author[a]{Yijiong Yu}

\affil[a]{Oregon State University\\
    \texttt{\{yuyiji, huazheng.wang\}@oregonstate.edu}
}

\author[a]{Huazheng Wang}
\author[b]{Shuai Yuan}
\author[b]{Ruilong Ren}
\author[b]{Ji Pei}

\affil[b]{DeepSolution\\
\texttt{research@deepsolution.chat}}

\begin{document}
\maketitle
\begin{abstract}
    Speculative Decoding (SD) accelerates low-concurrency LLM inference with a draft-then-verify paradigm.
    Mainstream methods, however, rely on multi-token prediction, which incurs compounding prediction difficulty and exposed draft latency.
    We propose Speculative Pipeline Decoding (SPD), which partitions the target LLM into $n$ pipeline stages so that $n$ tokens of a single sequence advance in parallel.
    To keep the pipeline saturated, a Pipeline Draft Module (PDM) aggregates multi-depth target features to predict the next token and runs concurrently with each pipeline step, yielding bounded prediction difficulty, higher acceptance, and hidden draft latency.
    Experiments show that SPD achieves higher theoretical and wall-clock speedup than EAGLE-3 at moderate pipeline width, while more aggressive widths still leave room for further gains.
    Our code is available at \href{https://github.com/yuyijiong/speculative_pipeline_decoding}{github}.

\end{abstract}

\section{Introduction}

The inference process of Large Language Models (LLMs) is bottlenecked by its autoregressive nature, which fundamentally restricts performance to memory bandwidth rather than compute capacity.
Speculative Decoding (SD) \citep{leviathan2023fast} has emerged as a prominent solution to mitigate this memory-bound latency.
Traditional SD employs a smaller, independent draft model to predict a sequence of future tokens, which are subsequently verified together by the target LLM.
Recent advancements, such as EAGLE \citep{li2024eagle}, bypass the need for an external draft model by appending a lightweight feature-extrapolation head to the target model, directly utilizing the LLM's internal hidden states to improve draft acceptance rates.

Despite these improvements, existing speculative decoding methods are fundamentally anchored to a \textbf{multi-token prediction paradigm}.
This paradigm inherently suffers from two critical, structural limitations that prevent further acceleration.

First, multi-token drafting suffers from \textbf{compounding feature drift}: after leaving the last target-verified token, later draft steps are increasingly conditioned on the draft module's own shallow hidden states rather than the target's, so acceptance decays sharply with draft length. This is why EAGLE-3 \citep{li2026eagle} typically keeps the draft length small (e.g., $k{=}3$) in practice.

Second, the multi-token paradigm introduces \textbf{exposed draft latency}.
Serial drafting leaves the target idle and offsets the verify-side speedup.
Recent methods such as DFlash \citep{chen2026dflash} reduce but do not fully remove this cost, and may trade off draft quality.

\begin{figure*}[!ht]
    \centering
    \includegraphics[width=1.9\columnwidth]{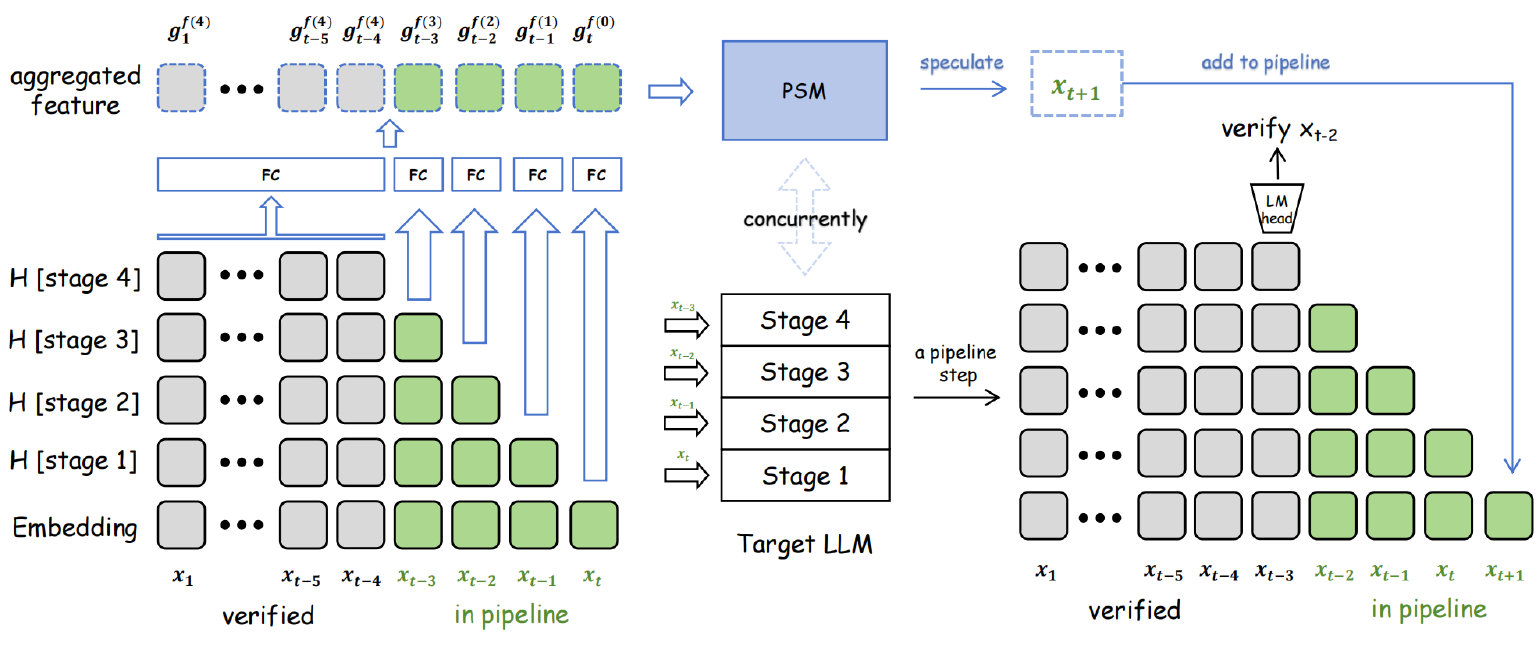}

    \caption{%
    Speculative Pipeline Decoding with $n{=}4$ stages.
    The target LLM is partitioned into four stages.
    At the beginning of a saturated cycle, in-flight tokens $x_{t-3},\ldots,x_t$ occupy the pipeline at varying depths, while $x_1,\ldots,x_{t-4}$ are already fully processed.
    For each token, hidden states at a fixed set of layer checkpoints (selected by how far that token has progressed) are concatenated and projected into one aggregated feature for the Pipeline Draft Module (PDM).
    The PDM drafts $\hat{x}_{t+1}$ concurrently with the target pipeline forward.
    After the forward, the oldest in-flight token $x_{t-3}$ exits and yields logits that verify $x_{t-2}$; if accepted, $\hat{x}_{t+1}$ enters stage~1 for the next cycle.
    }
    \label{fig:method}
\end{figure*}

To overcome these limitations, we propose \textbf{Speculative Pipeline Decoding (SPD)}.
We restructure the target LLM into an $n$-stage pipeline so that $n$ tokens of a single sequence advance at distinct depths concurrently.
This use of pipeline parallelism differs from the conventional setting in training and multi-request serving, where stages are filled with \emph{micro-batches} to hide pipeline bubbles and maximize throughput. SPD instead pipelines \emph{tokens of a single sequence} to reduce decode latency, and idle stages arise from a missing next token rather than a missing micro-batch.
A \textbf{Pipeline Draft Module (PDM)} therefore predicts a single next token each step to keep the pipeline saturated.
SPD has two key designs:

First, SPD uses \textbf{Multi-Depth Feature Aggregation} and does not condition on draft-generated features (Figure~\ref{fig:method}).
Each step predicts one token from target-LLM hidden states aggregated across pipeline depths, including partial states of in-flight tokens, so prediction stays in the target's feature space.
Difficulty is bounded by the fixed width $n$ rather than an open-ended draft horizon, yielding higher acceptance without choosing a draft length $k$.

Second, SPD \textbf{eliminates exposed serial drafting latency} by fully overlapping the PDM with the target pipeline step.
We shift the PDM forward so that it conditions on \emph{pre-step} features (intermediate states available before the current pipeline forward) rather than \emph{post-step} features (states after that forward), enabling true concurrency with the stage compute.
Pre-step features are slightly less informative, but multi-depth aggregation compensates, so a deeper PDM can still have its latency masked by the pipeline step.

We conduct extensive evaluations using Qwen3.5-4B and Qwen3.5-9B \citep{qwen3.5} across three representative benchmarks: MT-Bench \citep{zheng2023judging}, GSM8K \citep{cobbe2021training}, and HumanEval \citep{chen2021evaluating}.
Empirical results show that SPD surpasses EAGLE-3 \citep{li2026eagle} in wall-clock speedup across both model sizes at moderate pipeline width.
At larger widths (e.g., $n{=}16$), theoretical gains remain strong while measured speedup is currently tempered by per-cycle systems costs. This gap points to further headroom from systems-level optimizations rather than a limit of the decoding paradigm itself.

Our contributions are summarized as follows:
\begin{itemize}
    \item We formulate a pipeline-parallel speculative decoding paradigm that replaces unbounded multi-token draft horizons with a fixed pipeline width, structurally bounding draft difficulty while accelerating single-sequence decode.
    \item We show that conditioning a single-token drafter solely on multi-depth target features, and overlapping it with the pipeline step, yields higher acceptance while removing exposed serial drafting latency relative to drafting baselines.
    \item We provide distributed wall-clock evidence on multi-GPU hardware, and characterize when theory and practice diverge, pointing to concrete systems opportunities for closing the remaining gap at aggressive pipeline widths.
\end{itemize}

\begin{table*}[!t]
    \centering
    \caption{%
    Paradigm comparison.
    MTP = multi-token prediction (e.g., EAGLE-3); PP = pipeline parallelism.
    }
    \label{tab:paradigm}
    \small
    \begin{tabular}{lccc}
    \toprule
     & \textbf{MTP} & \textbf{PPSD} & \textbf{SPD (ours)} \\
    \midrule
    Target execution & single-path / batch verify & $n$-stage PP & $n$-stage PP \\
    Draft horizon & $k$ future tokens & 1 token/step & 1 token/step \\
    Draft features & target + draft HS & first-stage only & multi-depth target HS \\
    Draft timing & before verify (serial) & after PP step & with PP step \\
    Draft latency & often exposed & exposed & masked ($L_s{\approx}L/n{-}1$) \\
    \bottomrule
    \end{tabular}
    \end{table*}

\section{Related Work}

\textbf{Speculative Decoding.}
Standard speculative decoding \citep{leviathan2023fast} accelerates autoregressive generation by drafting several future tokens with a cheaper model and verifying them in one parallel target forward.
The systems intuition is that verification amortizes a single weight load over multiple candidates, raising arithmetic intensity on the target pass.
Self-speculative methods such as EAGLE and EAGLE-3 \citep{li2024eagle,li2026eagle} remove the external draft LM by attaching a lightweight head that extrapolates from the target's internal hidden states.
Despite higher acceptance than independent draft models, they remain multi-token predictors: later draft steps condition on the draft's own features, so acceptance decays with draft length and serial drafting still idles the target.
Table~\ref{tab:paradigm} contrasts this multi-token prediction (MTP) paradigm with PPSD and SPD.

\textbf{Parallel and Diffusion Drafting.}
A complementary line of work attacks draft latency itself.
P-EAGLE \citep{hui2026p} emits multiple draft tokens in parallel at quadratic training cost; Speculative Speculative Decoding \citep{kumar2026speculative} overlaps drafting with anticipated verification outcomes at the price of branching compute.
More recently, diffusion-style drafters push parallelism further inside the draft stage: DFlash \citep{chen2026dflash} replaces autoregressive drafting with a lightweight block-diffusion model that proposes an entire token block in one forward pass, while DSpark \citep{cheng2026dspark} pairs a parallel draft backbone with a lightweight sequential head and confidence-scheduled verification to curb suffix decay and verification waste under load.
These approaches improve how drafts are \emph{generated}, yet they still verify against a conventionally executed target.
SPD instead restructures the target's own execution; diffusion or semi-autoregressive drafters are therefore orthogonal ingredients that could, in principle, be composed with a pipeline-parallel target.

\textbf{Pipeline Drafting.}
Pipeline parallelism is standard in distributed training and serving, where micro-batches typically fill stages to hide bubbles and raise throughput.
Using the same stage partitioning to accelerate \emph{single-sequence} decode instead requires drafting the next token so that stages stay occupied without batching independent requests.
SpecPipe \citep{yin2025specpipe} combines pipelines with speculative decoding mainly as a system optimization for draft trees.
PPSD \citep{li2025pipeline} also uses pipeline parallelism for single-sequence decode, but drafts from first-stage features alone, so acceptance remains low, and still runs drafting after each pipeline step, leaving draft latency exposed.
SPD addresses both issues with multi-depth target aggregation and a pre-step schedule that overlaps the PDM with the pipeline step.

\section{Methodology}

We now detail Speculative Pipeline Decoding.
The design follows a simple narrative: pipeline-parallel target execution creates a need to supply the next token before earlier tokens finish (\S\ref{sec:pipeline}); filling that need without leaving the target's feature space motivates multi-depth aggregation (\S\ref{sec:aggregation}); an early, parallel draft schedule then hides draft cost behind each pipeline step (\S\ref{sec:simultaneous}); and streaming verification, phase-shifted from drafting by the pipeline width, keeps the schedule correct under accept/reject (\S\ref{sec:verify}).
Figure~\ref{fig:method} summarizes the end-to-end loop.

\subsection{Pipeline Execution and the Need for Drafting}
\label{sec:pipeline}
\label{sec:spec-overview}
Standard autoregressive decoding feeds one token through all $L$ layers before the next token can begin.
Pipeline parallelism instead partitions the target LLM into $n$ stages and advances $n$ tokens of the \emph{same} sequence at different depths in lockstep (Figure~\ref{fig:method}).
Each cycle then costs roughly one stage forward rather than a full model forward, which is the source of SPD's latency reduction.
This benefit, however, is available only if the pipeline stays full: every cycle must pop a finished token from the last stage \emph{and} push a new token into the first stage.
In single-sequence decoding that creates a circular dependency: the next token is not yet known because its predecessor has not produced final logits.
SPD breaks the circle by \textbf{drafting} the missing token while earlier tokens are still in flight.

Concretely, we attach a \textbf{Pipeline Draft Module (PDM)}: an $L_s$-layer Transformer decoder $f_\theta$ with an independent LM head $W_{\mathrm{lm}}$.
We fix the PDM to Qwen3-style standard attention even when the target (e.g., Qwen3.5) mixes attention types, avoiding linear-attention complexity in the drafter.
We deliberately predict only the \emph{single} next token per cycle, since multi-token drafting would reintroduce compounding feature drift inside the pipeline.

At step $t$, let $x_1,\ldots,x_t$ be the current sequence (verified prefix plus in-flight tokens).
Position $i$ carries a pipeline depth $k_i \in \{0,\ldots,n\}$ equal to how many stages $x_i$ has completed when drafting begins (\S\ref{sec:simultaneous}).
Rather than feeding raw token IDs, the PDM consumes a depth-conditioned feature sequence whose entries are derived from each token's already-computed target states (construction deferred to \S\ref{sec:aggregation}):
\begin{equation}
    \label{eq:input-seq}
    \begin{aligned}
    \mathcal{G}_t = &\bigl(g_1^{k_1},\, g_2^{k_2},\, \dots,\, g_t^{k_t}\bigr), \quad
    \mathbf{h}_t = \bigl[f_\theta(\mathcal{G}_t)\bigr]_t
    \\
    \hat{x}_{t+1} &\sim \mathrm{softmax}\bigl(W_{\mathrm{lm}}\mathbf{h}_t\bigr)
    \end{aligned}
\end{equation}
where $[\cdot]_t$ denotes the last-position representation.
Once the pipeline is saturated, verified positions all have depth $k{=}n$, while the $n$ in-flight tokens form a \emph{depth staircase} from almost-finished to newly embedded:
\begin{equation}
    \label{eq:steady-g}
    \mathcal{G}_t = \bigl(g_1^n, \ldots, g_{t-n}^n,\, g_{t-n+1}^{n-1}, \ldots, g_{t-1}^1,\, g_t^0\bigr).
\end{equation}
Eq.~\eqref{eq:steady-g} only records how far each token has progressed: larger $k$ means a richer (later-defined) feature, and $k{=}0$ is the newest token's embedding-level input.

\subsection{Multi-Depth Feature Aggregation}
\label{sec:aggregation}
We now define the features $g_i^{k}$ used in Eqs.~\eqref{eq:input-seq}--\eqref{eq:steady-g}.
PPSD uses only first-stage features, which is cheap but information-poor and degrades as $n$ grows (each stage becomes shallower).
EAGLE-3 already fuses multi-layer target hidden states on the verified prefix, but after leaving the last verified token it extrapolates from the draft module's own hidden states, which is precisely the path that causes compounding feature drift.
SPD takes a different route: for \emph{every} position, including in-flight tokens, use \textbf{as much of the target LLM as is already available}, and never condition on draft-generated features.

Let $H_t^l$ denote the hidden state of token $x_t$ after layer $l$ ($l{=}0$ is the embedding).
If $x_t$ has completed $k$ stages, the deepest available layer is $l_{\max}(k)=\min(L,\,k\cdot L/n)$.
Maintaining a separate fusion rule for every $k$ would be cumbersome, so we fix $m$ shared anchors $\mathcal{B}=[b_0,\ldots,b_{m-1}]$ with $b_0{=}0$ (e.g. for $L{=}32$ we use $[0,8,16,24,31]$).
A token at depth $k$ may consume every anchor that lies no deeper than $l_{\max}(k)$:
\begin{equation}
    \label{eq:depth-to-type}
    f(k) = \max \bigl\{\, i \;\big|\; b_i \le l_{\max}(k) \,\bigr\},
\end{equation}
\begin{equation}
    \label{eq:agg-main}
    g_t^k = \mathrm{FC}_{f(k)}\!\left(\mathrm{Concat}\!\left(H_t^{b_0}, \ldots, H_t^{b_{f(k)}}\right)\right).
\end{equation}
Thus $g_t^k$ is always built from $x_t$'s own target hidden states; the superscript $k$ selects \emph{which} anchors are eligible.
Verified positions in Eq.~\eqref{eq:steady-g} therefore share only the full-depth rule $k{=}n$ (the richest anchor set), while each $g_i^n$ remains distinct.
As a token advances through the pipeline, new anchors become eligible and $g_t^k$ grows strictly richer; at $k{=}0$ it reduces to a projection of the embedding.

Before the pipeline is full, only $a{<}n$ tokens are in flight.
Earlier positions, including the prefill prefix, are already fully processed, so they receive depth $k{=}n$; the trailing $a$ tokens form a shortened staircase:
\begin{equation}
    \label{eq:warmup-g}
    \mathcal{G}_t = \bigl(g_1^n, \ldots, g_{t-a}^n,\; g_{t-a+1}^{a-1}, \ldots, g_t^0\bigr).
\end{equation}
When $a{=}n$, this reduces to the steady-state layout in Eq.~\eqref{eq:steady-g}.

The incompleteness of any feature is therefore capped by the constant pipeline width $n$, not by an open-ended draft length.
This is why SPD need not tune a sensitive multi-token horizon: difficulty is structurally bounded, and drafting stays inside the target's feature geometry.
Training-time alignment with these layouts (sequence replication across aggregation types, simulated fill levels, and the structural attention mask) is detailed in Appendix~\ref{app:training}.

\subsection{Simultaneous Execution: Hiding Draft Latency Behind the Pipeline}
\label{sec:simultaneous}
Even an accurate drafter can offset its own gains if it runs serially with the target.
Classical speculative decoding must draft and then verify the \emph{same} candidates within one round, so the two stages are data-dependent and cannot overlap.
Under pipeline execution the two are phase-shifted by about $n$ steps (\S\ref{sec:verify}): a cycle verifies a token drafted earlier while the PDM proposes a token that will be checked only later.
With no intra-cycle draft$\rightarrow$verify dependency, drafting can proceed in parallel with the target forward.

To start with that forward, the PDM must use features already available at the beginning of the step.
We therefore condition it on \textbf{pre-step} features (intermediate hidden states present \emph{before} the current pipeline forward) rather than \textbf{post-step} features obtained after the forward completes.
Concretely, as soon as a newly drafted token enters stage~1, we launch the PDM on the current depth staircase: the newest token contributes only an embedding ($g^0$), and older in-flight tokens contribute their pre-step intermediates.
Relative to a post-step schedule, every in-flight position is one stage shallower (e.g., $g_t^0$ instead of $g_t^1$ for the newest token), so draft accuracy is slightly lower (e.g., ${\sim}5\%$ on Qwen3.5-4B).
The modest drop is more than repaid by latency masking (\S\ref{sec:ablation_io}).

True concurrency requires a dedicated draft rank: each of the $n$ stages occupies one GPU, and the PDM occupies an additional GPU that receives intermediate hidden states and runs aggregation plus decoding while stage workers compute.
This extra rank is the systems cost of keeping drafting off the critical path: trading one device for removing exposed serial draft latency.

For the overlap to be useful, drafting must finish no later than one stage forward.
Under equal per-layer cost this suggests $L_s \le L/n$; accounting for aggregation and the independent LM head, we adopt the practical rule $L_s \approx L/n{-}1$, which still allows a multi-layer drafter far deeper than a single-layer head.

\subsection{Streaming Verification and Rollback}
\label{sec:verify}
Lossless decoding still requires verifying drafted tokens against the target distribution.
Because the pipeline finishes one token per cycle, verification is \textbf{streaming}: we check one in-flight successor at a time rather than a long draft block.

We keep the indexing of Eq.~\eqref{eq:steady-g}: after a saturated step, $x_{1:t-n}$ is the verified prefix, $x_{t-n+1:t}$ are the in-flight tokens (oldest to newest), and the PDM is about to propose $\hat{x}_{t+1}$.
An in-flight successor becomes checkable only after its predecessor exits the last stage, so the token under verification was drafted about $n$ cycles earlier. This is the phase offset presupposed in \S\ref{sec:simultaneous}.

After each forward, the oldest in-flight token $x_{t-n+1}$ exits, is committed (it was accepted in the previous cycle), and yields $H_{t-n+1}^L$.
The target LM head forms $P(x_{t-n+2})$, which judges $x_{t-n+2}$.
Under greedy decoding we require an exact argmax match; under sampling we apply standard rejection sampling.

If $x_{t-n+2}$ is accepted, $\hat{x}_{t+1}$ enters stage~1 and the steady-state flow continues.
If it is rejected, we truncate target KV caches to length $t{-}n{+}1$ (discarding $x_{t-n+2},\ldots,x_t$), flush intermediate pipeline activations, and reseed stage~1 from $x_{t-n+2}\sim P(x_{t-n+2}\mid x_{1:t-n+1})$.

\section{Experiments}

\subsection{Experimental Setup}
\label{sec:setup}
\textbf{Training.}
Following EAGLE-3, we freeze all of the target LLM's parameters and train only the PDM.
Supervision is knowledge distillation: at every position the PDM predicts next-token logits from aggregated target features (Eq.~\eqref{eq:agg-main}), and we minimize KL divergence against the frozen teacher's (i.e., target LLM's) logits of the same token.
The PDM's LM head is initialized from the teacher's but with a smaller vocabulary to keep the drafter compact.
We mix common instruction-following datasets: ShareGPT-70k, UltraChat-200k \citep{ding2023enhancing}, SmolTalk \citep{allal2025smollm2smolgoesbig}, and SmolTalk-Chinese \citep{yu2025opencsgchinesecorpusseries}, filter out sequences longer than 2{,}048 tokens, and obtain about 1.2M samples.
Optimization uses learning rate $1\mathrm{e}{-}4$ with linear decay for a single epoch.
How training matches inference's varying pipeline occupancy is described in Appendix~\ref{app:training}.

\textbf{Evaluation.}
We evaluate on Qwen3.5-4B and Qwen3.5-9B \citep{qwen3.5} with thinking mode disabled (standard non-chain-of-thought decoding).
Following EAGLE-3, we report MT-Bench (chat), GSM8K (math), and HumanEval (code), with a maximum generation length of 512.
Sampling runs ($T{=}1$) use top-$k{=}50$ and top-$p{=}1.0$.
Baselines are EAGLE-3 \citep{li2026eagle} with draft length $k{\in}\{3,7,15\}$ and $L_s{=}1$, and PPSD \citep{li2025pipeline} with $n{\in}\{4,8,16\}$ and $L_s{=}1$.
Neither baseline hides draft latency: EAGLE-3 drafts serially before verify, and PPSD drafts after each pipeline step. Thus $L_s{>}1$ would multiply exposed draft cost and typically slow wall-clock time; following common practice we therefore keep their drafters at one layer.
For SPD, latency masking makes deeper drafters practical: we use uniform $L/n$ stage splits and select $L_s$ near $L/n - 1$.
Distributed wall-clock measurements run on multiple 8$\times$H20 GPU servers: SPD places each of the $n$ stages and the PDM on a separate rank ($n{+}1$ GPUs), so configurations such as $n{=}8$ (9 ranks) and $n{=}16$ (17 ranks) span more than one node; EAGLE-3 and the single-GPU autoregressive baseline use one H20 on the same cluster.
For wall-clock speedup, we run EAGLE-3 with the existing SGLang \citep{zheng2024sglang} serving stack, while PPSD and SPD use our own PyTorch inference framework implemented from scratch for pipeline-parallel speculative decoding.

\subsection{Metrics}
\label{sec:metrics}
Unless noted, all speedups are relative to standard autoregressive decoding of the same target LLM (single-GPU, batch size 1).
Speedup claims mix drafting quality with systems cost, so we report both theoretical and wall-clock speedup to show that SPD is stronger on the ideal upper bound and on real hardware.
Let $N$ be the number of generated tokens and $K$ the number of executed decoding steps (pipeline forwards for SPD; draft--verify rounds for EAGLE-3).
Rejections and flushes imply $K > N$ in general; we write $\alpha{=}N/K$ for the \emph{step acceptance ratio} (fraction of steps that ultimately commit a token).

Ignoring draft cost, one SPD cycle advances the equivalent of $n$ parallel stage tokens at $1/n$ of a full forward, yielding the \emph{oracle} (zero-draft-cost) theoretical speedup
\begin{equation}
    \label{eq:s0-spd}
    \mathcal{S}^0_{\mathrm{spd}} = \frac{N}{K \cdot (1/n)} = \alpha \cdot n,
\end{equation}
which coincides with the usual acceptance length when drafting is free.
When draft latency is \emph{not} hidden, we apply a \emph{draft-cost-adjusted} factor for EAGLE-3 (serial $k$-token drafting before verify) and PPSD (draft after each pipeline step):
\begin{equation}
    \label{eq:s-eagle}
    \begin{aligned}
    \mathcal{S}_{\mathrm{eagle}} &= \mathcal{S}^0_{\mathrm{eagle}} \cdot \frac{L}{L_s \cdot k + L}, \\
    \mathcal{S}_{\mathrm{ppsd}} &= \mathcal{S}^0_{\mathrm{ppsd}} \cdot \frac{L}{L_s \cdot n + L}.
    \end{aligned}
\end{equation}
Under the practical masking rule, SPD sets $\mathcal{S}_{\mathrm{spd}}=\mathcal{S}^0_{\mathrm{spd}}$.
Wall-clock speedup then asks whether these idealizations survive real hardware: we measure decode-phase tokens/s against the same single-GPU autoregressive baseline on the H20 cluster above (prefill excluded; batch size 1), and report the ratio as wall-clock speedup.

\subsection{Main Results and Analysis}
\label{sec:real_speedup}

Table~\ref{tab:speedup} reports \textbf{theoretical / wall-clock speedup} for EAGLE-3, PPSD, and SPD.
\input{tables/tab_speedup}

\textbf{Comparison with baselines.}
PPSD remains the weakest (often ${\lesssim}1.5\times$ wall-clock) due to shallow features and serial draft-after-pipeline execution.
SPD's best setting ($n{=}8$, $L_s{=}2$ or $3$) beats EAGLE-3 on overall wall-clock for both models and temperatures (e.g., $2.53\times$ vs.\ $1.97\times$ on 4B at $T{=}0$; $2.67\times$ vs.\ $2.54\times$ on 9B), with a larger margin under sampling and on HumanEval.

\begin{figure*}[t]
\centering
\includegraphics[width=0.95\textwidth]{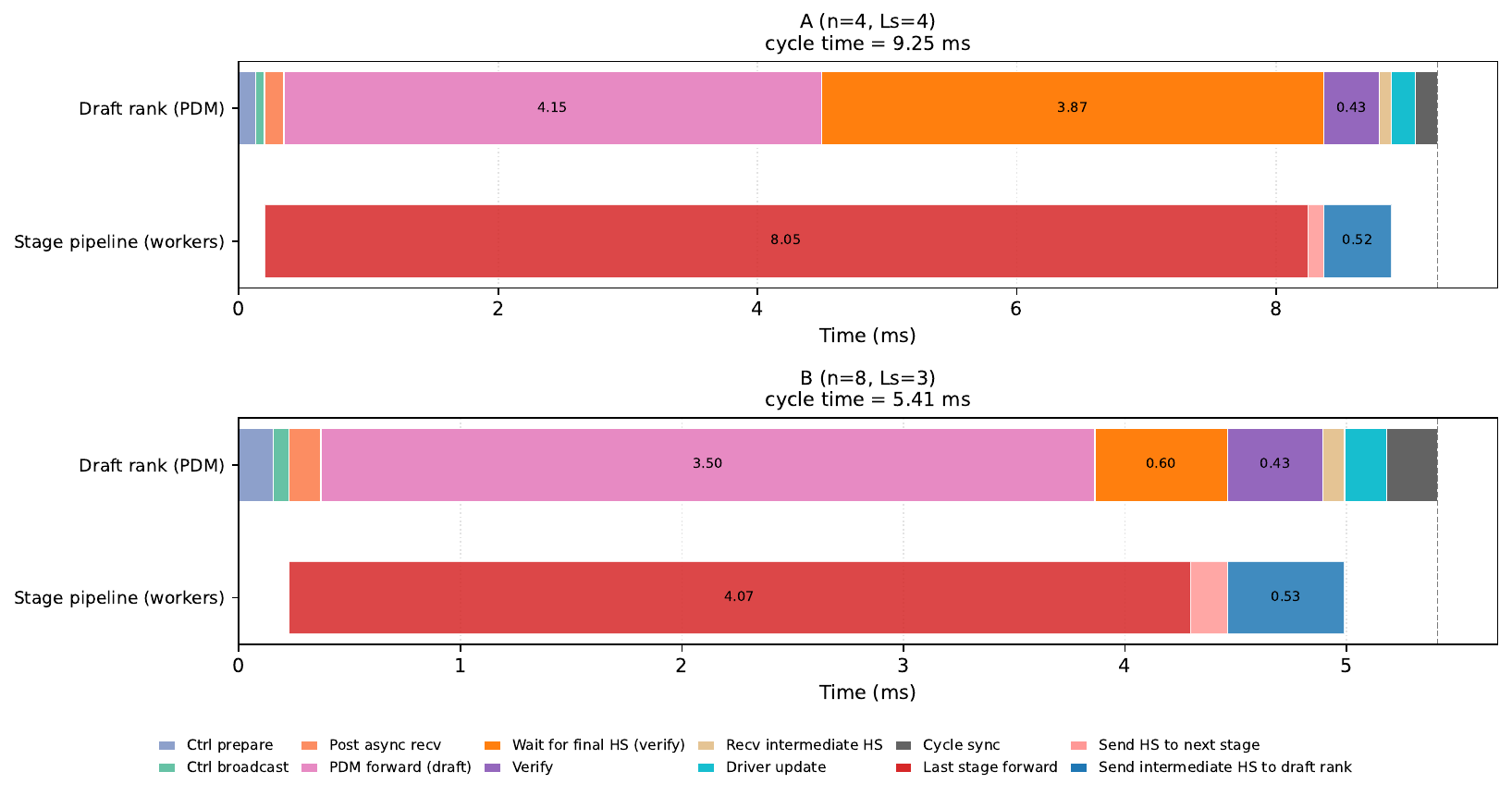}
\caption{%
Steady-state (pipeline full) per-cycle wall-time breakdown on Qwen3.5-4B for (A)~$n{=}4$, $L_s{=}4$ and (B)~$n{=}8$, $L_s{=}3$.
Drafting on the dedicated PDM GPU overlaps worker-side stage forward; fixed verification/transfer/control costs grow relatively as stages get shallower.
}
\label{fig:cycle_breakdown}
\end{figure*}

\textbf{Choosing the number of PDM layers $L_s$ at $n{=}8$.}
At fixed width $n{=}8$ we deliberately sweep $L_s\in\{2,3,4\}$ to expose the tension that motivates $L_s\approx L/n{-}1$.
Deeper PDMs improve draft quality: Table~\ref{tab:spec_layers} shows that raising $L_s$ from $1$ to $2$ lifts step acceptance substantially, with further but diminishing gains at $3$--$4$ layers.
At the same time, drafting must finish within one stage forward to stay masked.
Figure~\ref{fig:cycle_breakdown}(B) shows that at $L_s{=}3$ ($L/n{-}1$ for $L{=}32$), draft time is just slightly below stage forward time; one more layer would push drafting onto the critical path and re-expose latency.
Wall-clock in Table~\ref{tab:speedup} reflects both effects: $L_s{=}4$ can edge out theoretical speedup yet often loses measured throughput, while $L_s{=}2$ or $3$ is best in practice.
We therefore treat $L_s\approx L/n{-}1$ as an operating rule rather than a hard optimum: enough capacity for acceptance, but not so much that masking fails.

\input{tables/tab_spec_layers}

\textbf{Pipeline width and fixed per-cycle overhead.}
Profiling also explains why theory favors large $n$ while wall-clock does not.
Ideally, doubling $n$ halves stage compute and thus cycle time.
In Figure~\ref{fig:cycle_breakdown}, however, steady-state cycle wall only drops from $9.25$\,ms at $n{=}4$ to $5.41$\,ms at $n{=}8$, far from a $2\times$ reduction, because every cycle carries fixed costs (verification, communication/synchronization, etc.) that do not shrink with stage depth.
As $n$ grows, these fixed costs become a larger fraction of the cycle, so further widening yields less wall-clock return and eventually hurts (e.g., $n{=}16$ wall-clock collapses despite higher theoretical speedup).
Larger target models mitigate the same issue: longer per-stage forwards make fixed overhead relatively smaller, so measured speedup tracks theory more closely.
Empirically we therefore prefer moderate $n{\in}\{4,8\}$ over aggressive $n{=}16$, and expect SPD to be most attractive when the target LLM itself is large enough that stage computation dominates the cycle.

\subsection{Draft Trees}
\label{sec:draft_tree}
EAGLE-3 can sample multiple candidates per step into a \emph{draft tree} to raise acceptance; SPD admits the same branching.
We find that tree drafting (width=4) clearly improves acceptance and theoretical speedup relative to single-path drafting (Appendix~\ref{app:draft_tree}).
However, wall-clock speedup for draft trees is omitted because it requires complex system implementations such as branched KV caches, tree attention masks, and cross-stage synchronization, which we leave for future work.

\subsection{Ablation: Pre-step vs.\ Post-step Features}
\label{sec:ablation_io}

The schedule in \S\ref{sec:simultaneous} drafts from shallower \emph{pre-step} features so that drafting can start with the pipeline forward.
Waiting for deeper \emph{post-step} features raises the zero-draft-cost theoretical speedup (Table~\ref{tab:ablation_output}), especially at large $n$, but re-serializes drafting behind the target step.
Once that unhidden draft cost is counted, the draft-cost-adjusted theoretical speedup collapses (e.g., to $1.83$ at $n{=}16$).
The ablation supports the design choice: a modest accuracy drop is preferable to re-exposing draft latency.

\input{tables/tab_ablation_output}

\section{Conclusion}

We presented Speculative Pipeline Decoding, a shift from multi-token drafting to pipeline-parallel target execution.
By predicting one token at a time from multi-depth target features, SPD bounds draft difficulty by pipeline width; by overlapping a pre-step PDM with each pipeline step, it hides draft latency.
Distributed measurements show that moderate configurations ($n{=}8$, $L_s{\approx}L/n{-}1$) outperform EAGLE-3 in wall-clock speedup on Qwen3.5-4B and 9B, while overly large $n$ is limited by fixed per-cycle overhead.
We hope this perspective encourages a new paradigm of speculative decoding for more aggressive single-sequence speedups.

\section{Limitations}
\label{sec:limitations}

\textbf{Arithmetic intensity.}
Unlike conventional speculative decoding, which mainly raises per-device arithmetic intensity by verifying multiple candidates in one target forward, SPD accelerates a single sequence by using more GPUs in parallel and does not itself improve the compute-to-memory ratio on each device.
The two approaches are largely orthogonal, however: combining SPD with intensity-raising drafters (e.g., EAGLE-style or diffusion/semi-autoregressive methods) or with tensor parallelism is a promising route to further speedup.

\textbf{Hardware footprint.}
SPD requires $n$ stage ranks plus a dedicated PDM rank, so it uses more GPUs than single-device speculative decoding.
This cost is most acceptable precisely where large models are already multi-GPU deployed; on such targets, longer stage forwards also amortize fixed per-cycle overhead better, aligning practical speedup more closely with theory.
A related engineering direction is \emph{non-uniform} stage partitioning, which co-locates the PDM with a shallower first stage to keep the rank count within a single 4- or 8-GPU node. We leave this as exploratory future work.

\bibliography{custom2}
\appendix

\input{appendix}

\end{document}

%% file: tables/tab_speedup.tex
\begin{table*}[t!]
\centering
\caption{Single-path drafting on Qwen3.5-4B/9B ($L{=}32$). Format: theoretical / wall-clock speedup. MT = MT-Bench; HE = HumanEval.}
\label{tab:speedup}
\scriptsize
\setlength{\tabcolsep}{3.5pt}
\begin{tabular}{lcccccccc}
\toprule
& \multicolumn{4}{c}{\textbf{Qwen3.5-4B}} & \multicolumn{4}{c}{\textbf{Qwen3.5-9B}} \\
\cmidrule(lr){2-5} \cmidrule(lr){6-9}
\textbf{Method} & \textbf{Overall} & \textbf{MT} & \textbf{GSM8K} & \textbf{HE} & \textbf{Overall} & \textbf{MT} & \textbf{GSM8K} & \textbf{HE} \\
\midrule
\multicolumn{9}{l}{\textit{Temperature $T{=}0$}} \\
\midrule
EAGLE-3 ($k{=}3$) & 2.47 / 1.88 & 2.11 / 1.66 & 2.71 / 2.04 & 2.58 / 1.95 & 2.69 / 2.24 & 2.28 / 1.95 & 2.88 / 2.36 & 2.92 / 2.40 \\
EAGLE-3 ($k{=}7$) & 2.72 / 1.97 & 2.14 / 1.65 & 3.10 / 2.17 & 2.92 / 2.10 & 3.17 / 2.54 & 2.35 / 1.99 & 3.44 / \textbf{2.69} & 3.71 / 2.94 \\
EAGLE-3 ($k{=}15$) & 2.39 / 1.62 & 1.81 / 1.30 & 2.73 / 1.72 & 2.63 / 1.83 & 2.88 / 2.18 & 2.01 / 1.62 & 3.13 / 2.22 & 3.49 / 2.69 \\
\cmidrule(lr){1-9}
PPSD ($n{=}4$, $L_s{=}1$) & 1.54 / 1.28 & 1.41 / 1.17 & 1.52 / 1.26 & 1.70 / 1.41 & 1.90 / 1.53 & 1.57 / 1.28 & 1.85 / 1.49 & 2.27 / 1.82 \\
PPSD ($n{=}8$, $L_s{=}1$) & 1.68 / 1.25 & 1.38 / 1.04 & 1.60 / 1.20 & 2.05 / 1.52 & 1.96 / 1.44 & 1.51 / 1.13 & 1.82 / 1.34 & 2.55 / 1.85 \\
PPSD ($n{=}16$, $L_s{=}1$) & 1.40 / 0.89 & 1.18 / 0.75 & 1.36 / 0.86 & 1.65 / 1.04 & 1.56 / 0.98 & 1.23 / 0.78 & 1.48 / 0.92 & 1.98 / 1.23 \\
\cmidrule(lr){1-9}
Ours ($n{=}4$, $L_s{=}4$) & 2.64 / 2.27 & 2.19 / 1.90 & 2.61 / 2.24 & 3.11 / 2.65 & 2.70 / 2.32 & 2.24 / 1.94 & 2.68 / 2.30 & 3.19 / 2.73 \\
Ours ($n{=}8$, $L_s{=}4$) & 3.51 / 2.41 & 2.63 / 1.84 & 3.32 / 2.30 & 4.59 / 3.10 & 3.62 / 2.44 & 2.69 / 1.85 & 3.48 / 2.35 & 4.70 / 3.10 \\
Ours ($n{=}8$, $L_s{=}3$) & 3.39 / \textbf{2.53} & 2.55 / \textbf{1.93} & 3.22 / \textbf{2.43} & 4.40 / \textbf{3.23} & 3.61 / \textbf{2.67} & 2.65 / \textbf{2.01} & 3.43 / 2.55 & 4.74 / \textbf{3.44} \\
Ours ($n{=}8$, $L_s{=}2$) & 3.32 / 2.47 & 2.48 / 1.88 & 3.15 / 2.36 & 4.34 / 3.17 & 3.46 / 2.56 & 2.59 / 1.96 & 3.26 / 2.42 & 4.54 / 3.31 \\
Ours ($n{=}16$, $L_s{=}2$) & 3.70 / 1.43 & 2.59 / 1.04 & 3.35 / 1.32 & 5.17 / 1.94 & 3.90 / 1.49 & 2.71 / 1.08 & 3.52 / 1.37 & 5.48 / 2.02 \\
\midrule
\multicolumn{9}{l}{\textit{Temperature $T{=}1$}} \\
\midrule
EAGLE-3 ($k{=}3$) & 2.03 / 1.47 & 1.80 / 1.31 & 2.10 / 1.53 & 2.19 / 1.58 & 2.13 / 1.81 & 1.86 / 1.60 & 2.23 / 1.88 & 2.29 / 1.96 \\
EAGLE-3 ($k{=}7$) & 2.06 / 1.46 & 1.73 / 1.26 & 2.15 / 1.55 & 2.29 / 1.56 & 2.26 / 1.78 & 1.89 / 1.53 & 2.35 / 1.91 & 2.53 / 1.89 \\
EAGLE-3 ($k{=}15$) & 1.80 / 1.17 & 1.45 / 0.98 & 1.87 / 1.21 & 2.08 / 1.33 & 1.95 / 1.40 & 1.55 / 1.22 & 2.00 / 1.47 & 2.30 / 1.51 \\
\cmidrule(lr){1-9}
PPSD ($n{=}4$, $L_s{=}1$) & 1.39 / 1.06 & 1.30 / 0.99 & 1.37 / 1.05 & 1.50 / 1.14 & 1.71 / 1.27 & 1.44 / 1.08 & 1.67 / 1.24 & 2.03 / 1.49 \\
PPSD ($n{=}8$, $L_s{=}1$) & 1.44 / 0.95 & 1.27 / 0.84 & 1.40 / 0.92 & 1.66 / 1.09 & 1.68 / 1.10 & 1.36 / 0.90 & 1.57 / 1.03 & 2.12 / 1.37 \\
PPSD ($n{=}16$, $L_s{=}1$) & 1.20 / 0.65 & 1.06 / 0.58 & 1.19 / 0.64 & 1.33 / 0.72 & 1.32 / 0.72 & 1.10 / 0.61 & 1.27 / 0.69 & 1.60 / 0.86 \\
\cmidrule(lr){1-9}
Ours ($n{=}4$, $L_s{=}4$) & 2.48 / 1.95 & 2.09 / \textbf{1.67} & 2.49 / 1.96 & 2.85 / 2.21 & 2.57 / 2.01 & 2.13 / 1.72 & 2.54 / 1.99 & 3.03 / 2.33 \\
Ours ($n{=}8$, $L_s{=}4$) & 3.13 / 1.84 & 2.41 / 1.45 & 3.02 / 1.80 & 3.96 / 2.28 & 3.34 / 1.93 & 2.52 / 1.51 & 3.18 / 1.86 & 4.31 / 2.43 \\
Ours ($n{=}8$, $L_s{=}3$) & 3.05 / \textbf{2.05} & 2.40 / 1.66 & 2.91 / \textbf{1.99} & 3.84 / \textbf{2.50} & 3.33 / \textbf{2.21} & 2.50 / \textbf{1.73} & 3.16 / \textbf{2.13} & 4.32 / \textbf{2.78} \\
Ours ($n{=}8$, $L_s{=}2$) & 2.96 / 2.03 & 2.29 / 1.62 & 2.86 / \textbf{1.99} & 3.73 / 2.47 & 3.16 / 2.15 & 2.36 / 1.68 & 3.01 / 2.07 & 4.11 / 2.69 \\
Ours ($n{=}16$, $L_s{=}2$) & 3.19 / 1.06 & 2.37 / 0.81 & 2.96 / 1.00 & 4.24 / 1.37 & 3.46 / 1.13 & 2.45 / 0.83 & 3.11 / 1.03 & 4.83 / 1.53 \\
\bottomrule
\end{tabular}
\end{table*}

%% file: tables/tab_spec_layers.tex
\begin{table}[!t]
\centering
\caption{Effect of PDM layer count $L_s$ on step acceptance ratio $\alpha{=}N/K$ (Qwen3.5-4B, single-path drafting; averaged over MT-Bench, GSM8K, and HumanEval).}
\label{tab:spec_layers}
\small
\begin{tabular}{cccc}
\toprule
$n$ & $L_s$ & $T{=}0$ & $T{=}1$ \\
\midrule
8 & 1 & 0.2977 & 0.2708 \\
8 & 2 & 0.3589 & 0.3276 \\
8 & 3 & 0.3721 & 0.3396 \\
8 & 4 & 0.3799 & 0.3435 \\
4 & 1 & 0.5352 & 0.5007 \\
4 & 4 & 0.6119 & 0.5752 \\
\bottomrule
\end{tabular}
\end{table}

%% file: tables/tab_ablation_output.tex
\begin{table}[!t]
    \centering
    \caption{Qwen3.5-4B when the PDM uses \emph{post-step} features (breaks parallelism). Format: zero-draft-cost theoretical speedup / draft-cost-adjusted theoretical speedup, averaged over MT-Bench, GSM8K, and HumanEval.}
    \label{tab:ablation_output}
    \resizebox{\columnwidth}{!}{%
        \begin{tabular}{lcccc}
            \toprule
            \textbf{Method}      & \textbf{T=0} & \textbf{T=1} & \textbf{T=0 + tree} & \textbf{T=1 + tree} \\
            \midrule
            Ours ($n=4, L_s=2$)  & 2.61 / 2.09       & 2.50 / 2.00       & 2.70 / 2.16       & 2.96 / 2.36       \\
            Ours ($n=8, L_s=2$)  & 3.31 / 2.20       & 3.09 / 2.06       & 3.40 / 2.27       & 4.10 / 2.73       \\
            Ours ($n=16, L_s=2$) & 3.66 / 1.83       & 3.28 / 1.64       & 3.69 / 1.84       & 4.78 / 2.39       \\
            \bottomrule
        \end{tabular}%
    }
\end{table}

%% file: appendix.tex
\section{Training Alignment with Pre-step Pipeline Drafting}
\label{app:training}
\label{sec:dynamic-fill}

The main text defines multi-depth aggregation (Eqs.~\eqref{eq:depth-to-type}--\eqref{eq:agg-main}) and the steady / warm-up feature layouts (Eqs.~\eqref{eq:steady-g},~\eqref{eq:warmup-g}).
At inference the PDM always consumes \emph{pre-step} features, so the newest in-flight token contributes only $g^0$, and the depth staircase depends on how many tokens currently occupy the pipeline.
Training must therefore expose the same layout variation; this appendix describes the three mechanisms that achieve that alignment: sequence replication across aggregation types, randomized pipeline-fill simulation, and the structural attention mask.
We keep the Qwen3.5 anchors fixed at $\mathcal{B}{=}[0,8,16,24,31]$ ($m{=}5$); exact choices matter little (${\sim}2\%$ acceptance) as long as shallow, middle, and deep layers are represented, while dropping the embedding anchor ($b_0{=}0$) prevents stable convergence.

\subsection{Sequence Replication for Multi-Depth Supervision}

A single fused feature per token cannot represent coexisting aggregation types or the cross-depth visibility that the staircase requires.
We therefore \textbf{replicate} every position into $m$ contiguous blocks (deepest to shallowest), expanding sequence length from $N$ to $mN$.
The frozen target is run once per sequence; each block applies Eq.~\eqref{eq:agg-main} with the corresponding anchor prefix.
The LM head reads only the shallowest block ($g^0$) at each position, matching the pre-step inference schedule.
Completed full-depth features $g_i^n$ remain valid across steps and may be KV-cached; any $g_i^k$ with $k{<}n$ is refreshed on the next advance, so caching appends only the trailing $(n{+}1)$ refreshed features each step.

\subsection{Warm-up Layouts and Simulated Pipeline Fill}
\label{app:warmup}

When the pipeline is saturated ($a{=}n$), the PDM sees the full depth staircase of Eq.~\eqref{eq:steady-g}.
During warm-up, and briefly after a flush, only $a{<}n$ tokens are in flight, so the layout collapses to the shortened staircase of Eq.~\eqref{eq:warmup-g}: the prefix uses depth $k{=}n$, and only the trailing $a$ positions form a staircase.
These two families of layouts are not interchangeable; a PDM trained only on the saturated pattern would be miscalibrated whenever the pipeline is partially filled.
We therefore train under a \emph{simulated} fill level that randomly reproduces both regimes.
For each sequence, with probability $0.5$ we set $\mathrm{fill}{=}n$ (saturated); otherwise $\mathrm{fill}$ is drawn uniformly from $\{1,\dots,n{-}1\}$ (warm-up).
The structural mask below then forces every attention pattern to obey the corresponding layout, so the PDM learns to draft under both depth staircases.

\subsection{Structural Attention Mask}

Every replicated block participates in attention; the mask alone encodes which cross-depth interactions are legal under the sampled fill.
A query at time $t$ and depth $d$ attending to key $T$ requires key depth $d'{=}\min(n,\, d + t - T)$ in block $f(d')$.
If $t{-}T$ exceeds $\mathrm{fill}$, the key is redirected to the deepest block, matching the depth-maximized prefix of warm-up Eq.~\eqref{eq:warmup-g}.
Causal order is preserved ($T{\le}t$).
Together with sequence replication, this mask makes training occupancy match the pre-step inference layouts without running an actual pipeline during training.

\section{Draft-Tree Theoretical Speedup}
\label{app:draft_tree}

As noted in \S\ref{sec:draft_tree}, SPD can expand each draft step into a width-$4$ draft tree.
Table~\ref{tab:draft4} reports the resulting \textbf{theoretical} speedup only (wall-clock for branched drafting is not yet implemented).
Relative to single-path drafting in the main tables, trees raise acceptance and thus $\mathcal{S}$ for SPD (e.g., overall $3.05{\rightarrow}3.84$ on Qwen3.5-4B at $T{=}1$ for $n{=}8$, $L_s{=}3$).
For EAGLE-3 and PPSD the numbers use the draft-cost-adjusted $\mathcal{S}$ of Eq.~\eqref{eq:s-eagle}; for SPD, latency masking keeps $\mathcal{S}{=}\mathcal{S}^0$.

\input{tables/tab_draft4}

Under trees, EAGLE-3 often leads on greedy decoding for the larger model (overall and MT-Bench / GSM8K at $T{=}0$), while SPD is stronger under sampling and on HumanEval.
PPSD remains weakest: trees do not fix shallow first-stage features and can hurt when incorrect branches crowd out shorter correct paths.
These figures characterize draft quality under branching, not deployable wall-clock speedups.

%% file: tables/tab_draft4.tex
\begin{table*}[t!]
\centering
\caption{Theoretical speedup with draft trees on Qwen3.5-4B/9B. MT = MT-Bench; HE = HumanEval.}
\label{tab:draft4}
\scriptsize
\setlength{\tabcolsep}{3.5pt}
\begin{tabular}{lcccccccc}
\toprule
& \multicolumn{4}{c}{\textbf{Qwen3.5-4B}} & \multicolumn{4}{c}{\textbf{Qwen3.5-9B}} \\
\cmidrule(lr){2-5} \cmidrule(lr){6-9}
\textbf{Method} & \textbf{Overall} & \textbf{MT} & \textbf{GSM8K} & \textbf{HE} & \textbf{Overall} & \textbf{MT} & \textbf{GSM8K} & \textbf{HE} \\
\midrule
\multicolumn{9}{l}{\textit{Temperature $T{=}0$}} \\
\midrule
EAGLE-3 ($k{=}3$) & 2.75 & 2.15 & 3.14 & 2.97 & 3.16 & 2.82 & 3.36 & 3.30 \\
EAGLE-3 ($k{=}7$) & 3.61 & \textbf{2.86} & \textbf{4.15} & 3.81 & \textbf{4.14} & \textbf{3.19} & \textbf{4.58} & 4.66 \\
EAGLE-3 ($k{=}15$) & 3.33 & 2.55 & 3.80 & 3.66 & 4.07 & 2.86 & 4.37 & 4.97 \\
\cmidrule(lr){1-9}
PPSD ($n{=}4$, $L_s{=}1$) & 1.54 & 1.41 & 1.56 & 1.70 & 1.80 & 1.54 & 1.88 & 2.22 \\
PPSD ($n{=}8$, $L_s{=}1$) & 1.49 & 1.28 & 1.53 & 1.83 & 1.66 & 1.36 & 1.71 & 2.27 \\
PPSD ($n{=}16$, $L_s{=}1$) & 1.18 & 1.04 & 1.22 & 1.38 & 1.28 & 1.08 & 1.31 & 1.64 \\
\cmidrule(lr){1-9}
Ours ($n{=}4$, $L_s{=}4$) & 2.70 & 2.28 & 2.68 & 3.13 & 2.77 & 2.32 & 2.76 & 3.23 \\
Ours ($n{=}8$, $L_s{=}4$) & 3.59 & 2.69 & 3.45 & 4.64 & 3.72 & 2.77 & 3.60 & 4.80 \\
Ours ($n{=}8$, $L_s{=}3$) & 3.46 & 2.60 & 3.37 & 4.43 & 3.69 & 2.72 & 3.55 & 4.79 \\
Ours ($n{=}8$, $L_s{=}2$) & 3.40 & 2.54 & 3.26 & 4.40 & 3.55 & 2.65 & 3.38 & 4.62 \\
Ours ($n{=}16$, $L_s{=}2$) & \textbf{3.75} & 2.62 & 3.42 & \textbf{5.22} & 4.00 & 2.78 & 3.64 & \textbf{5.58} \\
\midrule
\multicolumn{9}{l}{\textit{Temperature $T{=}1$}} \\
\midrule
EAGLE-3 ($k{=}3$) & 2.63 & 2.38 & 2.76 & 2.75 & 2.63 & 2.39 & 2.68 & 2.83 \\
EAGLE-3 ($k{=}7$) & 2.79 & 2.32 & 2.84 & 3.20 & 2.99 & 2.47 & 3.15 & 3.34 \\
EAGLE-3 ($k{=}15$) & 2.44 & 2.01 & 2.52 & 2.78 & 2.71 & 2.22 & 2.79 & 3.12 \\
\cmidrule(lr){1-9}
PPSD ($n{=}4$, $L_s{=}1$) & 1.62 & 1.50 & 1.65 & 1.73 & 1.89 & 1.63 & 1.97 & 2.22 \\
PPSD ($n{=}8$, $L_s{=}1$) & 1.67 & 1.44 & 1.74 & 1.97 & 1.88 & 1.55 & 1.93 & 2.45 \\
PPSD ($n{=}16$, $L_s{=}1$) & 1.35 & 1.16 & 1.41 & 1.56 & 1.47 & 1.22 & 1.52 & 1.84 \\
\cmidrule(lr){1-9}
Ours ($n{=}4$, $L_s{=}4$) & 2.85 & 2.53 & 2.79 & 3.25 & 2.88 & 2.56 & 2.83 & 3.26 \\
Ours ($n{=}8$, $L_s{=}4$) & 3.93 & 3.22 & 3.69 & 4.89 & 4.08 & 3.28 & 3.88 & 5.10 \\
Ours ($n{=}8$, $L_s{=}3$) & 3.84 & 3.15 & 3.66 & 4.72 & 4.03 & 3.21 & 3.81 & 5.07 \\
Ours ($n{=}8$, $L_s{=}2$) & 3.74 & 3.06 & 3.53 & 4.64 & 3.88 & 3.14 & 3.65 & 4.84 \\
Ours ($n{=}16$, $L_s{=}2$) & \textbf{4.30} & \textbf{3.38} & \textbf{3.80} & \textbf{5.73} & \textbf{4.53} & \textbf{3.48} & \textbf{3.96} & \textbf{6.16} \\
\bottomrule
\end{tabular}
\end{table*}

%% file: custom2.bib
@article{zheng2023judging,
  title={Judging llm-as-a-judge with mt-bench and chatbot arena},
  author={Zheng, Lianmin and Chiang, Wei-Lin and Sheng, Ying and Zhuang, Siyuan and Wu, Zhanghao and Zhuang, Yonghao and Lin, Zi and Li, Zhuohan and Li, Dacheng and Xing, Eric and others},
  journal={Advances in neural information processing systems},
  volume={36},
  pages={46595--46623},
  year={2023}
}

@article{cobbe2021training,
  title={Training verifiers to solve math word problems, 2021},
  author={Cobbe, Karl and Kosaraju, Vineet and Bavarian, Mohammad and Chen, Mark and Jun, Heewoo and Kaiser, Lukasz and Plappert, Matthias and Tworek, Jerry and Hilton, Jacob and Nakano, Reiichiro and others},
  journal={URL https://arxiv.org/abs/2110.14168},
  volume={9},
  year={2021}
}

@misc{chen2021evaluating,
      title={Evaluating Large Language Models Trained on Code},
      author={Mark Chen and Jerry Tworek and Heewoo Jun and Qiming Yuan and Henrique Ponde de Oliveira Pinto and Jared Kaplan and Harri Edwards and Yuri Burda and Nicholas Joseph and Greg Brockman and Alex Ray and Raul Puri and Gretchen Krueger and Michael Petrov and Heidy Khlaaf and Girish Sastry and Pamela Mishkin and Brooke Chan and Scott Gray and Nick Ryder and Mikhail Pavlov and Alethea Power and Lukasz Kaiser and Mohammad Bavarian and Clemens Winter and Philippe Tillet and Felipe Petroski Such and Dave Cummings and Matthias Plappert and Fotios Chantzis and Elizabeth Barnes and Ariel Herbert-Voss and William Hebgen Guss and Alex Nichol and Alex Paino and Nikolas Tezak and Jie Tang and Igor Babuschkin and Suchir Balaji and Shantanu Jain and William Saunders and Christopher Hesse and Andrew N. Carr and Jan Leike and Josh Achiam and Vedant Misra and Evan Morikawa and Alec Radford and Matthew Knight and Miles Brundage and Mira Murati and Katie Mayer and Peter Welinder and Bob McGrew and Dario Amodei and Sam McCandlish and Ilya Sutskever and Wojciech Zaremba},
      year={2021},
      eprint={2107.03374},
      archivePrefix={arXiv},
      primaryClass={cs.LG}
}

@misc{ding2023enhancing,
      title={Enhancing Chat Language Models by Scaling High-quality Instructional Conversations}, 
      author={Ning Ding and Yulin Chen and Bokai Xu and Yujia Qin and Zhi Zheng and Shengding Hu and Zhiyuan Liu and Maosong Sun and Bowen Zhou},
      year={2023},
      eprint={2305.14233},
      archivePrefix={arXiv},
      primaryClass={cs.CL}
}

@misc{allal2025smollm2smolgoesbig,
      title={SmolLM2: When Smol Goes Big -- Data-Centric Training of a Small Language Model}, 
      author={Loubna Ben Allal and Anton Lozhkov and Elie Bakouch and Gabriel Martín Blázquez and Guilherme Penedo and Lewis Tunstall and Andrés Marafioti and Hynek Kydlíček and Agustín Piqueres Lajarín and Vaibhav Srivastav and Joshua Lochner and Caleb Fahlgren and Xuan-Son Nguyen and Clémentine Fourrier and Ben Burtenshaw and Hugo Larcher and Haojun Zhao and Cyril Zakka and Mathieu Morlon and Colin Raffel and Leandro von Werra and Thomas Wolf},
      year={2025},
      eprint={2502.02737},
      archivePrefix={arXiv},
      primaryClass={cs.CL},
      url={https://arxiv.org/abs/2502.02737}, 
}

@misc{yu2025opencsgchinesecorpusseries,
      title={OpenCSG Chinese Corpus: A Series of High-quality Chinese Datasets for LLM Training}, 
      author={Yijiong Yu and Ziyun Dai and Zekun Wang and Wei Wang and Ran Chen and Ji Pei},
      year={2025},
      eprint={2501.08197},
      archivePrefix={arXiv},
      primaryClass={cs.CL},
      url={https://arxiv.org/abs/2501.08197}, 
}

@article{li2026eagle,
  title={Eagle-3: Scaling up inference acceleration of large language models via training-time test},
  author={Li, Yuhui and Wei, Fangyun and Zhang, Chao and Zhang, Hongyang},
  journal={Advances in Neural Information Processing Systems},
  volume={38},
  pages={136737--136756},
  year={2026}
}

@inproceedings{li2024eagle,
  title={EAGLE: speculative sampling requires rethinking feature uncertainty},
  author={Li, Yuhui and Wei, Fangyun and Zhang, Chao and Zhang, Hongyang},
  booktitle={Proceedings of the 41st International Conference on Machine Learning},
  pages={28935--28948},
  year={2024}
}

@article{hui2026p,
  title={P-EAGLE: Parallel-Drafting EAGLE with Scalable Training},
  author={Hui, Mude and Huang, Xin and Salas, Jaime Campos and Sun, Yue and Pemberton, Nathan and Song, Xiang and Khetan, Ashish and Karypis, George},
  journal={arXiv preprint arXiv:2602.01469},
  year={2026}
}

@inproceedings{leviathan2023fast,
  title={Fast inference from transformers via speculative decoding},
  author={Leviathan, Yaniv and Kalman, Matan and Matias, Yossi},
  booktitle={International Conference on Machine Learning},
  pages={19274--19286},
  year={2023},
  organization={PMLR}
}

@article{li2025pipeline,
  title={Pipeline Parallelism is All You Need for Optimized Early-Exit Based Self-Speculative Decoding},
  author={Li, Ruanjun and Liu, Ziheng and Shi, Yuanming and Shao, Jiawei and Zhang, Chi and Li, Xuelong},
  journal={arXiv preprint arXiv:2509.19368},
  year={2025}
}

@article{kumar2026speculative,
  title={Speculative speculative decoding},
  author={Kumar, Tanishq and Dao, Tri and May, Avner},
  journal={arXiv preprint arXiv:2603.03251},
  year={2026}
}

@misc{qwen3.5,
    title  = {{Qwen3.5}: Towards Native Multimodal Agents},
    author = {{Qwen Team}},
    month  = {February},
    year   = {2026},
    url    = {https://qwen.ai/blog?id=qwen3.5}
}

@article{zheng2024sglang,
  title={Sglang: Efficient execution of structured language model programs},
  author={Zheng, Lianmin and Yin, Liangsheng and Xie, Zhiqiang and Sun, Chuyue and Huang, Jeff and Yu, Cody H and Cao, Shiyi and Kozyrakis, Christos and Stoica, Ion and Gonzalez, Joseph E and others},
  journal={Advances in neural information processing systems},
  volume={37},
  pages={62557--62583},
  year={2024}
}

@article{yin2025specpipe,
  title={SpecPipe: Accelerating Pipeline Parallelism-based LLM Inference with Speculative Decoding},
  author={Yin, Haofei and Xiao, Mengbai and Li, Tinghong and Zhang, Xiao and Yu, Dongxiao and Zhang, Guanghui},
  journal={arXiv preprint arXiv:2504.04104},
  year={2025}
}

@misc{chen2026dflash,
  title={{DFlash}: Block Diffusion for Flash Speculative Decoding},
  author={Chen, Jian and Liang, Yesheng and Liu, Zhijian},
  year={2026},
  eprint={2602.06036},
  archivePrefix={arXiv},
  primaryClass={cs.CL},
  url={https://arxiv.org/abs/2602.06036}
}

@misc{cheng2026dspark,
  title={{DSpark}: Confidence-Scheduled Speculative Decoding with Semi-Autoregressive Generation},
  author={Cheng, Xin and Yu, Xingkai and Shao, Chenze and Li, Jiashi and Xiong, Yunfan and Qian, Yi and Zhu, Jiaqi and Ma, Shirong and Zhang, Xiaokang and Ye, Jiasheng and others},
  year={2026},
  eprint={2607.05147},
  archivePrefix={arXiv},
  primaryClass={cs.CL},
  url={https://arxiv.org/abs/2607.05147}
}
